\documentclass[11pt]{article}

\usepackage[preprint]{acl}

\usepackage{times}
\usepackage{latexsym}
\usepackage[utf8]{inputenc}
\usepackage[T1]{fontenc}
\usepackage{booktabs}
\usepackage{amsmath}
\usepackage{amsfonts}
\usepackage{nicefrac}
\usepackage{microtype}
\usepackage{inconsolata}
\usepackage{xcolor}
\usepackage{graphicx}
\usepackage{capt-of}
\usepackage{placeins}
\usepackage{float}

\newcommand{\sage}{\textsc{SAGE}}
\newcommand{\socialevo}{\textsc{SocialEvo}}
\newcommand{\selfevo}{\textsc{SelfEvo}}
\newcommand{\teg}{\ensuremath{\mathrm{TEG}}}
\newcommand{\E}{\mathbb{E}}

\title{\sage: A Quantitative Evaluation of Socialized Evolution in Agent Ecosystems}

\author{
    Linyue Pan\textsuperscript{1},
    Yaoming Zhu\textsuperscript{2},
    Lin Qiu\textsuperscript{2},
    Xuezhi Cao\textsuperscript{2},
    Xunliang Cai\textsuperscript{2}\\
    \textsuperscript{1}Tsinghua University, China
    \textsuperscript{2}Meituan, China\\
    \texttt{ply24@mails.tsinghua.edu.cn}, \texttt{ym-zhu@outlook.com}
}

\begin{document}

\maketitle

\begin{abstract}
Self-improving language agents are typically evaluated in isolation: an agent attempts a task, receives feedback, and iteratively refines its own behavior.
Yet agents increasingly operate alongside peers whose strategies and outcomes are publicly visible.
This raises an under-studied question: \emph{when does shared experience produce improvements that self-improvement alone cannot achieve?}
We introduce \textbf{\sage{}}---\textbf{S}ocial \textbf{A}gent \textbf{G}roup \textbf{E}volution---an evaluation framework that compares two compute-matched conditions: \textbf{\socialevo{}}, where agents from five distinct model families co-evolve with access to all peers' histories; and \textbf{\selfevo{}}, where each agent receives the same number of task attempts but sees only its own past, which is conventional in self-improving agent studies.
We instantiate \sage{} in three arenas---open-ended ML research, long-horizon economic planning, and strategic multiplayer play---evaluated across multiple evolutionary rounds.
We find that group history is not a universal amplifier: the strongest agent does not exceed its self-evolution ceiling.
However, agents that plateau under self-improvement can achieve significant breakthroughs when peer experience is available.
In competitive settings, counterfactual controls reveal that agents improve generally rather than developing opponent-specific strategies.
Across different forms of shared history, filtered peer traces and reflective summaries often outperform raw logs, indicating that social gains depend on abstraction rather than exposure volume.
These findings reveal that peer-history gains are agent-specific, arena-dependent, and contingent on the capacity to abstract transferable knowledge from public traces.
\end{abstract}

\section{Introduction}

Language agents are usually evaluated as solitary problem solvers: an agent receives a task, executes a trajectory, and is scored.
Recent work on self-improving agents makes this view iterative rather than static: agents can revise prompts, update memories, refine tool-use policies, and improve from their own past attempts \citep{madaan2023selfrefine,shinn2023reflexion,wang2023voyager,robeyns2025selfimproving,zhang2025darwin}.
Iteration alone still misses a feature of real agent ecosystems.
In research platforms, coding competitions, and multiplayer games, agents learn from more than their own histories: they observe peers' successes and failures, consult shared leaderboards, and face competitors whose strategies continually co-evolve.

This paper asks whether such public experience produces a measurable form of \emph{socialized evolution}: improvements that peer exposure adds beyond self-improvement.
We introduce \textbf{\sage{}} (\textbf{S}ocial \textbf{A}gent \textbf{G}roup \textbf{E}volution), an evaluation framework for measuring this gap.
\sage{} compares two conditions.
In \textbf{\socialevo{}}, agents from five distinct model families co-evolve in a shared ecosystem and can access all peers' histories.
In \textbf{\selfevo{}}, each focal agent receives the same number of task attempts but observes only its own past, matching the conventional setup in self-improving agent studies.

The utility of public experience depends on the interaction setting.
Research in multi-agent reinforcement learning and emerging LLM societies separates two mechanisms relevant here: \emph{cooperative knowledge transfer}, where agents absorb peers' discoveries to master complex environments \citep{ndousse2021emergent,zhuge2023mindstorms}; and \emph{adversarial co-evolution}, where agents iteratively adapt to and counter opponent strategies \citep{baker2019emergent,vinyals2019grandmaster,li2024gtbench}.
We use three arenas: open-ended ML research (MLR-Bench) and long-horizon economic strategy (DrugWars) test cooperative social learning, while the adversarial board game Splendor tests whether group history produces opponent-specific adaptation under competitive pressure.
We also vary the \emph{form} of shared history, including raw trajectories, reflective summaries, and filtered leaderboard signals, to identify which representations agents can learn from and which mainly add noise.

\begin{figure*}[!t]
\centering
\includegraphics[width=\textwidth]{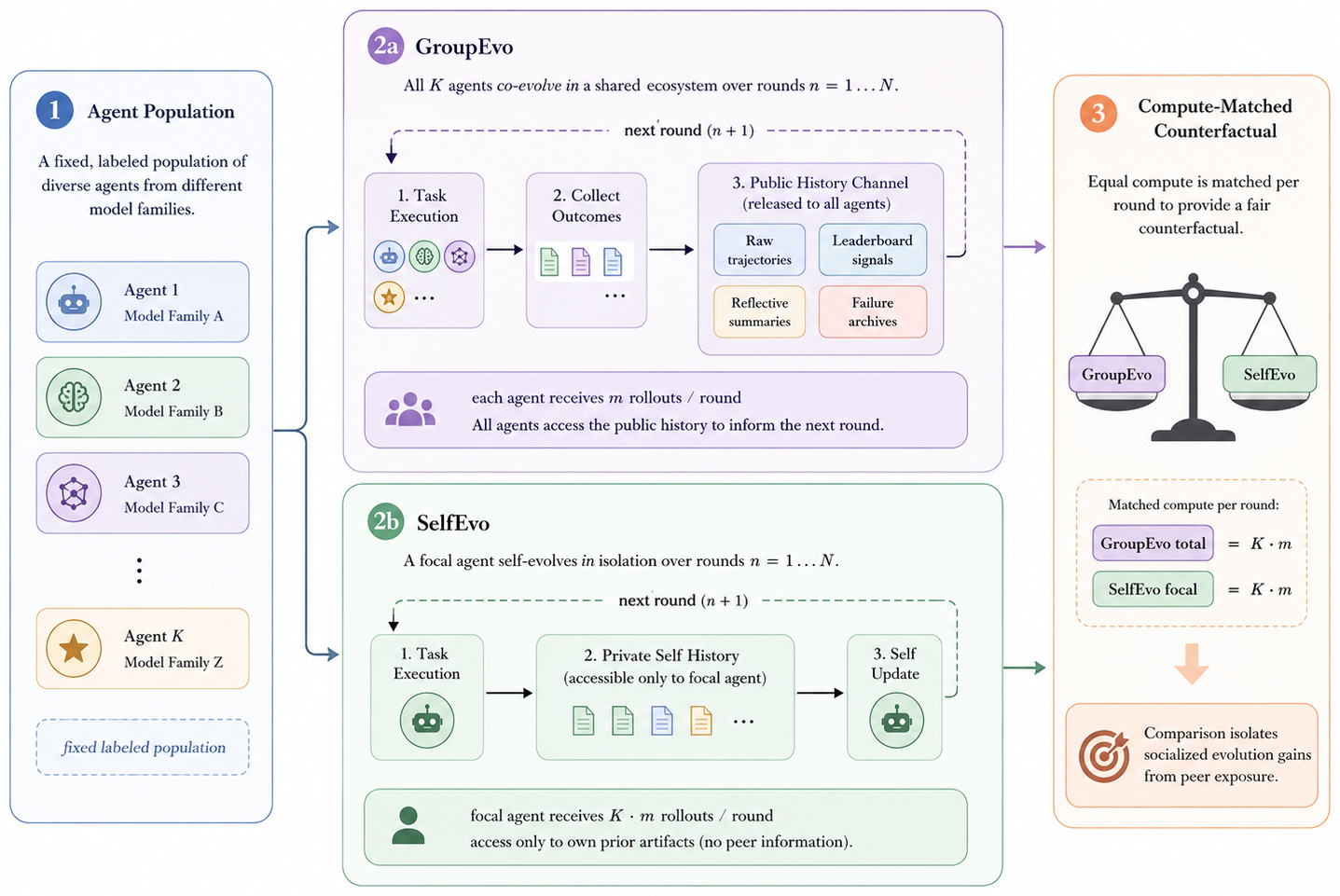}
\caption{\sage{} evaluation framework.
A fixed population of labeled agents enters two compute-matched regimes: \socialevo{}, where all agents co-evolve with access to the public history channel, and \selfevo{}, where a focal agent evolves in isolation with an equal per-round rollout budget but only private history.
The comparison isolates gains attributable to peer exposure rather than additional test-time compute.}
\label{fig:sage-framework}
\end{figure*}

Our contributions are:
\begin{itemize}
    \item We propose a compute-matched design for socialized agent evolution.
    \sage{} compares \socialevo{} and \selfevo{} under matched rollout budgets, isolating peer-history effects from generic test-time iteration.

    \item We evaluate cooperative and adversarial social learning in heterogeneous arenas.
    We instantiate \sage{} in open-ended ML research, long-horizon economic planning, and strategic multiplayer play, covering both cooperative knowledge transfer and adversarial co-evolution \citep{baker2019emergent,li2024gtbench}.

    \item We report agent-level evidence on socialized evolvability.
    Social gains are not determined by static strength: matched-compute gains appear for selected agents, while the same public channel can hurt agents that fail to filter or reuse peer traces.

    \item We add diagnostics for opponent-specific adaptation and history representation.
    We introduce \teg{}-style analysis and model-swap shadow tests for ranked play, and ablate public-history representations to distinguish useful social information from noisy context.
\end{itemize}

\section{The \sage{} Framework}

Figure~\ref{fig:sage-framework} presents the fixed agent population, paired \socialevo{} and \selfevo{} evolution loops, and matched-compute comparison.

\subsection{Socialized Evolution}

Let $\mathcal{A}=\{a_1,\ldots,a_K\}$ denote a fixed and labeled population of agents, and let $n\in\{1,\ldots,N\}$ index evolution rounds.
At each round, an arena runs each agent under task-specific rules, records outcomes, and may release artifacts to later rounds.
These artifacts can include raw trajectories, leaderboard positions, evaluator feedback, summaries, failure records, or ranked match records.
The object measured by \sage{} is not a single score, but a population of trajectories:
\[
    \left\{ \mathcal{H}_i^1, y_i^1, \ldots, \mathcal{H}_i^N, y_i^N \right\}_{i=1}^{K},
\]
where $\mathcal{H}_i^n$ is the history visible to agent $i$ before round $n$, and $y_i^n$ is the arena-native outcome.

The main intervention is the \emph{history channel}.
A history channel defines the visibility scope, content granularity, and representation format of shared artifacts.
Different representations of the same events, such as raw logs, reflective summaries, failure archives, and leaderboards, can lead to different learning behavior, so the channel itself is an experimental variable.

\subsection{Compute-Matched Counterfactuals}

\sage{} compares two evolution regimes.

\paragraph{\socialevo{} condition.}
The $K$ agents co-evolve in a shared ecosystem under \socialevo{}.
Each agent receives $m$ rollouts per round and may access the public-history channel.
The total rollout budget per round is $K m$.

\paragraph{\selfevo{} baseline.}
A focal agent evolves alone.
It receives $K m$ rollouts per round, matching the total budget used by the full group, but it can access only its own prior artifacts.
The self condition captures what repeated practice, self-reflection, and extra sampling can achieve without peer exposure.

Let $\mathcal{G}_i^n$ and $\mathcal{S}_i^n$ denote the round-$n$ performance of agent $i$ under the \socialevo{} condition and the \selfevo{} baseline.
The compute-matched comparison is necessary: without it, social gains may simply reflect more attempts rather than better use of peer experience.

\subsection{History Representations}

We treat the representation of public history as an experimental variable.
The DrugWars history-mode ablation instantiates five concrete channels:
\begin{itemize}
    \item \textbf{Full history:} agents may inspect all previous public traces and leaderboards.
    \item \textbf{No history:} agents receive no previous public history.
    \item \textbf{Leaderboard-only:} agents see only previous rankings and scores.
    \item \textbf{Top-1 trace:} agents see only the previous round champion trace.
    \item \textbf{Summary:} agents see only a generated public summary of the previous round.
\end{itemize}
These modes separate whether social learning relies on raw demonstrations, reputation signals, exemplar imitation, or compressed public memory.

\section{Experimental Setup}

\subsection{Arenas}

\paragraph{MLR-Bench.}
MLR-Bench evaluates agents on open-ended machine-learning research tasks \citep{chen2025mlrbench}.
Agents generate ideas, write proposals, conduct experiments, and produce final papers.
The score aggregates research-quality judgments over multiple stages.
This arena tests open-ended scientific reasoning and makes proposals, code traces, experimental outcomes, and judge feedback available for evolution.

\paragraph{DrugWars.}
DrugWars is a classic turn-based trading game in which a player starts with cash and debt, uses stochastic price movement over time, and tries to maximize end-of-horizon net worth while surviving adverse events \citep{m4cs2020drugwars,drugwars2026reloaded}.
Our environment is a deterministic, tool-driven adaptation of the open-source \texttt{M4cs/drugwars} implementation \citep{m4cs2020drugwars}.
We build the acting agent with Google's Agent Development Kit (ADK) \citep{google2026adk}.
Agents repeatedly make trading decisions, manage debt, and maximize liquidation value.
DrugWars tests tool use, capital allocation, risk control, and adaptation to market trajectories.
High-scoring runs often reveal reusable policies, making DrugWars a testbed for distinguishing strategy abstraction from exact trace copying.

\paragraph{Splendor.}
Splendor is a classic two-to-four-player resource-management board game in which players collect gem tokens, purchase development cards, attract nobles, and race to prestige points \citep{spacecowboys2026splendor}.
We implement Splendor as an executable game environment with a Google ADK agent interface \citep{google2026adk}.
Agents play Swiss-style matches over multiple rounds.
We treat public match histories as social experience and use final league rank, derived from Swiss standings, as the main performance metric.

\begin{figure*}[!t]
\centering
\includegraphics[width=\textwidth]{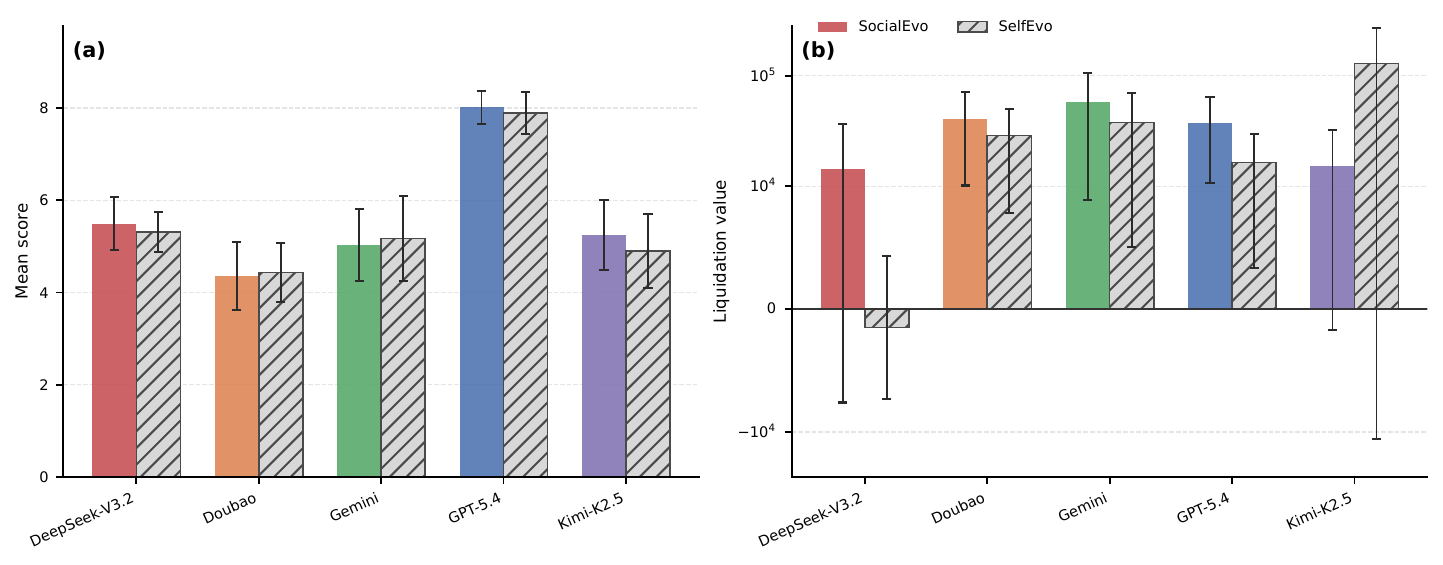}
\caption{RQ1 absolute performance under \socialevo{} and compute-matched \selfevo{} over post-initial evolution rounds.
Panel (a) reports MLR-Bench scores.
Panel (b) reports DrugWars liquidation values on a symlog scale.}
\label{fig:rq1-main}
\end{figure*}

\begin{figure*}[!t]
\centering
\includegraphics[width=\textwidth]{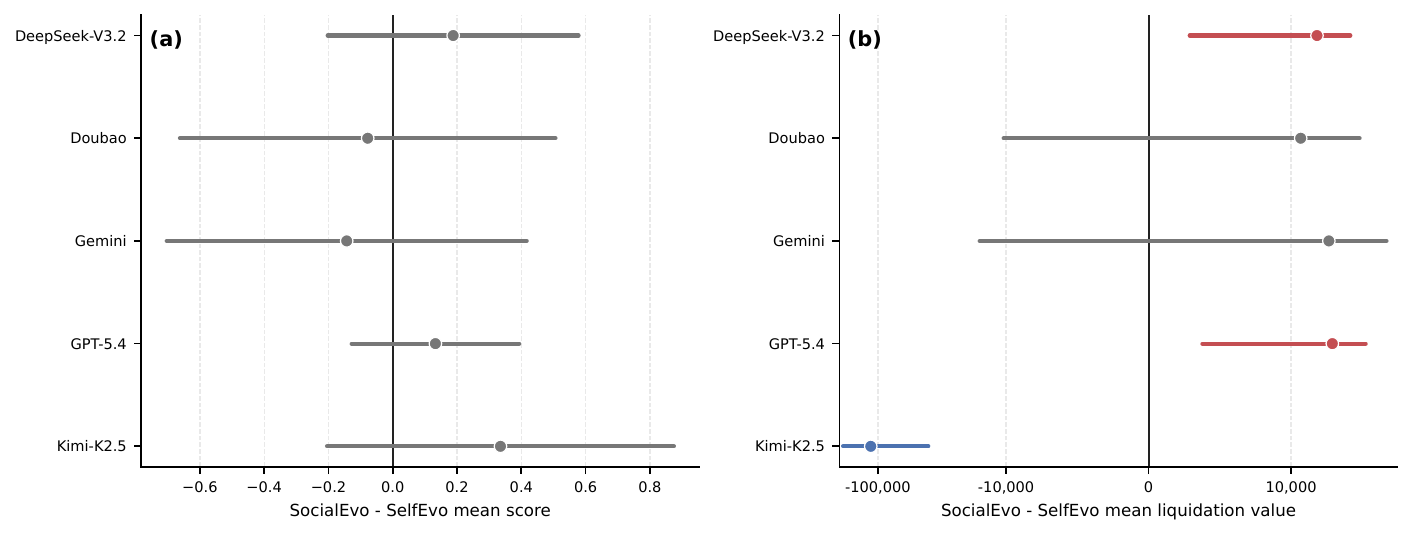}
\caption{RQ2 \socialevo{} effects with paired confidence intervals over post-initial evolution rounds.
Panel (a) reports MLR-Bench score differences.
Panel (b) reports DrugWars liquidation-value differences.
Red intervals are positive and do not cross zero, blue intervals are negative and do not cross zero, and gray intervals cross zero.}
\label{fig:rq2-forest}
\end{figure*}

\subsection{Agents and Rounds}

The main experiments use five agents from distinct model families.
We cite the corresponding public technical reports, model cards, or official release notes where available:
\texttt{DeepSeek-V3.2} \citep{deepseekai2025deepseekv32}, \texttt{doubao-seed-2-0-pro-260215} \citep{bytedanceseed2026seed20}, \texttt{gemini-3-flash-preview} \citep{googledeepmind2025gemini3flash}, \texttt{gpt-5.4} \citep{openai2026gpt54}, and \texttt{kimi-k2.5} \citep{moonshotai2026kimi}.
Unless otherwise stated, each arena runs for $N=16$ rounds with $m=1$ group rollout per agent per round.
The \selfevo{} baseline allocates the same total rollout budget to each focal agent.

\section{Results}

\subsection{RQ1: \socialevo{} Benefits Are Agent-Dependent}

RQ1 asks which agents benefit from \socialevo{} under compute-matched \selfevo{}.
For MLR-Bench and DrugWars, we compute per-round means and compare the post-initial evolution rounds, where round 1 is the no-history starting point.
The descriptive effect is
\[
    \Delta_i = \E[\mathcal{G}_i^{2:16}] - \E[\mathcal{S}_i^{2:16}],
\]
with confidence intervals estimated from paired per-round differences.
Figure~\ref{fig:rq1-main} compares the absolute \socialevo{} and \selfevo{} means for each agent, and Appendix Table~\ref{tab:rq1-main} reports the corresponding numeric summary.

\begin{figure*}[!t]
\centering
\includegraphics[width=\textwidth]{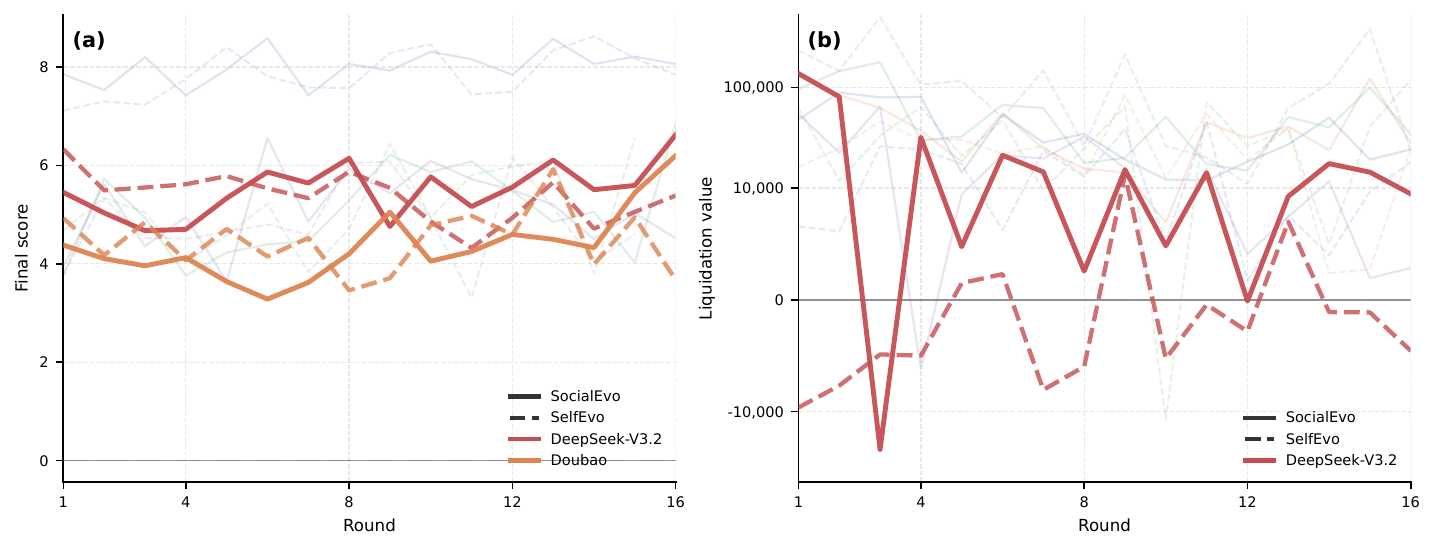}
\caption{RQ2 evolution curves under \socialevo{} and \selfevo{}.
Panel (a) reports MLR-Bench and highlights DeepSeek-V3.2 and Doubao.
Panel (b) reports DrugWars and highlights DeepSeek-V3.2.
Solid lines denote \socialevo{} means and dashed lines denote \selfevo{} means; non-highlighted agents are faded as background context.}
\label{fig:rq2-evolution}
\end{figure*}

The first pattern is that \socialevo{} is not a population-level free lunch.
Different models have different \socialevo{} ability: some can absorb peer histories as usable evidence, while others treat the same histories as noise or misleading demonstrations.
In MLR-Bench, the group and self curves remain close for every agent, including the top absolute performer.
One likely reason is that open-ended research tasks already give high \selfevo{} baselines enough internal feedback, leaving little room for peer traces to create a stable additional gain.
DrugWars behaves differently: public histories often expose reusable trading heuristics, but they also create opportunities for negative transfer.
Because \selfevo{} receives the same total rollout budget, these contrasts cannot be explained by giving the group more attempts.
They are better understood as selective strategy transfer: peer exposure helps only when the agent can identify which public traces encode portable decisions and ignore traces that are irrelevant or brittle.

\subsection{RQ2: Excess \socialevo{} Gains Are Not the Same as Static Strength}

RQ2 asks whether agents obtain excess socialized gains beyond what their \selfevo{} strength would predict.
Figure~\ref{fig:rq2-forest} shows \socialevo{} minus \selfevo{} effects with paired confidence intervals over the same post-initial rounds, and Figure~\ref{fig:rq2-evolution} shows the round-by-round trajectories.

The forest plot turns the directional comparison into a mixed result.
MLR-Bench shows no confirmed agent-level advantage, whereas DrugWars contains both confirmed positive and confirmed negative effects.
The same public channel can be a source of strategy discovery for some agents and a source of distraction for another.
The highest \selfevo{} DrugWars trajectory is also the one most harmed by \socialevo{}, so socialized evolvability cannot be reduced to static strength or final self-performance.
The more plausible explanation is an interaction between exploration frontier and absorption capacity.
When self-exploration is narrow, peer traces can add strategies the agent would not sample alone; when the agent already finds good strategies or cannot reliably filter low-quality traces, the same channel can dilute or derail its policy.
\socialevo{} measures a second-order capability: solving the task and deciding which pieces of public experience deserve to change one's own behavior.
The evolution curves add a temporal check on this interpretation.
They show whether a forest-plot effect is a persistent separation between \socialevo{} and \selfevo{} or merely a late-round fluctuation.
In DrugWars, the highlighted DeepSeek trajectory stays above its self baseline through much of the evolution process, which is stronger evidence of socialized learning than a single final-round comparison.
By contrast, the MLR-Bench curves mostly track their self baselines, matching the forest plot's conclusion that peer exposure does not create a stable extra margin there.
The curves therefore support the main claim of RQ2: SocialEvo ability differs across models and arenas, and stable gains require sustained use of public experience rather than one-off lucky rollouts.

\begin{figure*}[!t]
\centering
\includegraphics[width=\textwidth]{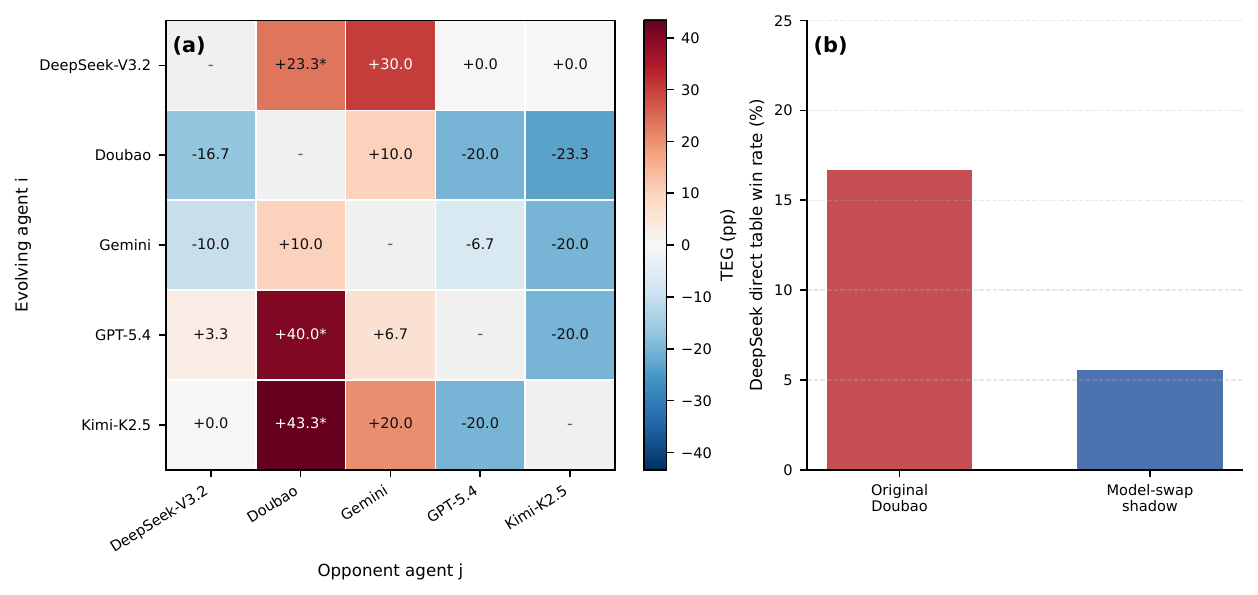}
\caption{RQ3 targeted evolution in Splendor.
Panel (a) reports $\teg_{i,j}$ in percentage points, where rows are evolving agents and columns are opponents.
Panel (b) reports the round-16 original-live-model repeat and pooled model-swap shadow conditions for the selected DeepSeek-to-Doubao pair.
The shadow condition tests whether DeepSeek's advantage drops after the live model behind the Doubao public label is replaced.}
\label{fig:rq3-splendor}
\end{figure*}

\subsection{RQ3: Opponent-Specific Adaptation in Ranked Play}

RQ3 asks whether ranked \socialevo{} produces targeted gains against particular opponents.
For Splendor, we define the targeted evolution gain from agent $i$ to opponent $j$ as
\[
    \teg_{i,j}
    =
    \left(\mathcal{G}_{i,j}^N - \mathcal{G}_{i,j}^1\right)
    -
    \left(\mathcal{S}_{i,j}^N - \mathcal{S}_{i,j}^1\right).
\]
Here $\mathcal{G}_{i,j}^n$ is the \socialevo{} win rate that $i$ ranks above $j$ in round $n$.
Splendor score ties are assigned average ranks, and a pairwise win is counted only when $i$ has a strictly better rank than $j$.
The current \selfevo{} runs are homogeneous same-model leagues, so $\mathcal{S}_{i,j}^n$ is estimated from independent \selfevo{} rank distributions and serves as a generic-progress baseline rather than a direct head-to-head self match.
Figure~\ref{fig:rq3-splendor} reports tie-aware Splendor \teg{} values from the five \socialevo{} repeats and homogeneous \selfevo{} baselines.
The notable structure is not a uniform rise in competitive ability: the cells that pass the exploratory gate all target Doubao.
This concentration points to opponent-conditioned adaptation rather than a uniform gain against every opponent.
The preselected DeepSeek-to-Doubao pair remains positive under the tie-aware reanalysis, making it a suitable target for a stricter identity-control test.
We run the confirmatory model-swap shadow test on this preselected pair: the public label and public history remain Doubao, while the live model behind that label is replaced.
The replacement runs are pooled into one shadow condition rather than analyzed by replacement model, and the original-live-model runs are repeated under the same replay protocol.
If DeepSeek had learned a Doubao-specific behavioral model, the shadow bar would drop relative to the original-live-model bar.
We report the pooled clean shadow contrast
\[
\begin{aligned}
    \mathrm{ShadowDrop}_{i,j}
    &= p_{\mathrm{orig}} - p_{\mathrm{swap}},\\
    p_{\mathrm{orig}}
    &= \widehat{\Pr}[i \succ j_{\mathrm{label}} \mid j_{\mathrm{live}}],\\
    p_{\mathrm{swap}}
    &= \widehat{\Pr}[i \succ j_{\mathrm{label}} \mid \mathrm{swap}_{\mathrm{live}}].
\end{aligned}
\]
Here $j_{\mathrm{label}}$ denotes the unchanged public identity and $\mathrm{swap}_{\mathrm{live}}$ pools all clean replacement runs.
After tied tables are corrected to shared average ranks, the model-swap bar drops.
For this preselected pair, the pattern is consistent with a Doubao-targeted strategy rather than generic league practice.
The shadow test matters here because it separates learning about a public identity from generic improvement: the advantage falls when the same label and history no longer correspond to the live Doubao model.

\subsection{RQ4: Public Social Information as the Intervention}

\begin{figure*}[!t]
\centering
\begin{minipage}[t]{0.48\textwidth}
\centering
\captionof{table}{RQ4 DrugWars history-mode ablation by agent.
Cells report mean liquidation values in thousands over 16 budget-clean rounds per mode.}
\label{tab:rq4-history-agent}
\scriptsize
\setlength{\tabcolsep}{3pt}
\resizebox{\linewidth}{!}{%
\begin{tabular}{lrrrrr}
\toprule
History mode & DeepSeek & Doubao & Gemini & GPT-5.4 & Kimi \\
\midrule
Full & 8.7 & 18.8 & 92.9 & 46.5 & 26.3 \\
No history & 15.5 & 15.5 & 48.3 & 15.4 & 46.4 \\
Leaderboard & 18.3 & 72.0 & 68.2 & 18.0 & 40.7 \\
Top-1 trace & 6.5 & 52.0 & 126.7 & 38.7 & 53.1 \\
Summary & 17.1 & 37.4 & 74.0 & 16.1 & 20.6 \\
\bottomrule
\end{tabular}}
\end{minipage}
\hfill
\begin{minipage}[t]{0.48\textwidth}
\centering
\vspace{0.45\baselineskip}
\includegraphics[width=\linewidth,height=0.43\linewidth,keepaspectratio]{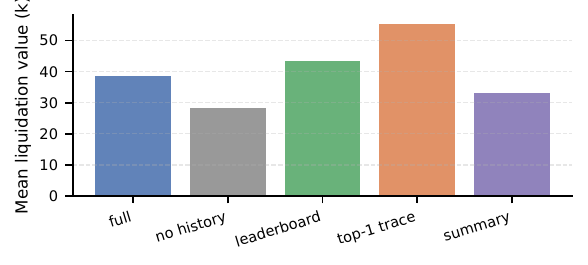}
\captionof{figure}{RQ4 overall mean liquidation value by history mode, averaged over 80 agent-round observations per mode.}
\label{fig:rq4-history}
\end{minipage}
\end{figure*}

RQ4 asks which public signals are enough for \socialevo{} gains, and whether the representation of public information explains the results better than its quantity.
Table~\ref{tab:rq4-history-agent} and Figure~\ref{fig:rq4-history} summarize the budget-clean DrugWars history-mode ablation.
The table reports per-agent means, and the figure reports the corresponding overall mean for each history mode.

In this ablation, representation explains more than raw information volume.
The highest-mean modes are not the ones that expose all traces; they are the modes that pre-filter social information into either a salient exemplar or a reputation signal.
This matches the interpretation from RQ2: agents do not simply benefit from more public context, because they must first solve a filtering problem.
Top-1 trace reduces the search problem to imitation of a visible high performer, while leaderboard-only gives weaker but still useful pressure by telling agents which behaviors were rewarded without exposing every action.
Full history can contain the same useful demonstrations, but it also adds distractors, stale failures, and a larger context burden.
The summary condition shows the opposite risk: compression can remove concrete action details and leave advice that is too abstract to execute.
\socialevo{} is also a social-information design problem.
The best channel is not the one that exposes the most history, but the one that makes public experience easiest to filter, abstract, and transfer.

\section{Related Work}

\paragraph{Self-improving agents.}
A growing literature studies agents that improve through self-feedback, memory, or iterative refinement \citep{madaan2023selfrefine,shinn2023reflexion,park2023generative,wang2023voyager}.
Recent work extends this idea to scaffold-level modification, reusable skills, and open-ended self-improvement \citep{zhao2023expel,robeyns2025selfimproving,zhang2025darwin,novikov2025alphaevolve,yuksekgonul2026learning}.
These methods show that self-history can improve agents.
\sage{} treats them as the baseline and asks what peer history adds beyond matched \selfevo{}.

\paragraph{Peer learning and social evolution.}
CATArena studies iterative tournament-style evaluation and peer learning for code agents \citep{fu2025catarena}.
Weng et al. treat a group rather than an individual as the unit of open-ended self-improvement and emphasize experience sharing in coding environments \citep{weng2026group}.
Other multi-agent systems study collaboration, debate, and role-structured execution \citep{li2023camel,wu2023autogen,du2024multiagent,wang2024rethinking,li2025rethinkingmoa}.
\sage{} differs by using compute-matched counterfactuals, heterogeneous non-code arenas, and diagnostics that distinguish general social gains from static task strength and brittle trace copying.

\paragraph{Memory and history representation.}
Agent performance depends on how history is stored, compressed, and retrieved, not merely on whether it exists.
Reflexion, Generative Agents, and Voyager use reflective memory or skill libraries to stabilize long-horizon behavior \citep{shinn2023reflexion,park2023generative,wang2023voyager}.
Recent memory benchmarks further separate raw recall from experience reuse and long-horizon adaptation \citep{maharana2024locomo,wu2024longmemeval,tan2025reflectivememory,xiong2025memorymanagement,wei2025evomemory,xie2026memevobench}.
\sage{} extends this concern from individual memory to public social memory.

\paragraph{Ranked and environment-centered evaluation.}
Recent LLM-agent benchmarks evaluate social deduction, negotiation, strategic interaction, and research workflows \citep{brown2022cicero,cheng2023avalonbench,chan2024negotiationtom,smith2025concordia,abdulhai2026har,lupu2025decrypto,belle2025agentsofchange,huang2024mlagentbench,chan2025mlebench,wijk2025rebench,nathani2025mlgym,chen2025mlrbench}.
Adaptive benchmark work such as Population Based Training, Dynabench, The Ladder, and MLRC-Bench shows that evolving populations and public leaderboards can change what benchmark progress measures \citep{jaderberg2017population,kiela2021dynabench,blum2015ladder,zhang2025mlrcbench}.
\sage{} follows this population-level view, but focuses on how public history changes agent adaptation trajectories.

\section{Discussion}

\subsection{Exposure Versus Learning}

A recurring failure mode is confusing exposure with learning.
An agent can read a peer trajectory and still fail to identify the transferable principle, copy a surface action that only works for one seed, or overweight a low-quality peer because the trace is recent or verbose.
The cross-round results show this distinction empirically: the same \socialevo{} channel helps some DrugWars agents but hurts Kimi-K2.5.
Social-learning agents need filtering, abstraction, and transfer mechanisms rather than mere access to more public artifacts.

\subsection{History Mode Does Not Mean More Is Better}

The DrugWars history ablation shows that larger raw context is not automatically superior.
Top-1 trace and leaderboard-only conditions are competitive with, and in this run stronger than, full raw history.
The ablation favors designing public history around the form and selectivity of social information rather than the amount alone.
Selected exemplars, reputation signals, and compressed summaries impose different abstraction burdens on the agent, and the highest-scoring channel is the one the agent can actually convert into action.

\section{Conclusion}

\sage{} asks a question that static benchmarks cannot answer: how do agents evolve when experience becomes public?
By comparing the \socialevo{} condition against the compute-matched \selfevo{} baseline, the framework separates peer-exposure gains from generic iterative improvement.
The results do not give an unqualified endorsement of multi-agent exposure.
Public history does not reliably improve every agent, but the cross-round DrugWars confidence intervals are above zero for DeepSeek-V3.2 and GPT-5.4 and below zero for Kimi-K2.5.
The Splendor shadow test shows that DeepSeek learned a Doubao-targeted strategy in the selected pair.
These mixed outcomes make strategy abstraction and public-information design explicit parts of the evaluation.
For shared agent ecosystems, evaluation needs to measure whether agents can filter, abstract, and reuse public experience.

\section{Limitations}

This work has two scope limitations.
First, the experiments use a fixed population of five mainstream model families.
This design keeps comparisons controlled, but the distribution of socialized gains may change with smaller models, open-source agents, or specialized scaffolds.
Second, our budget constrains the amount of public history that can be evaluated within each agent context.
As a result, the present study tests bounded shared-history settings rather than arbitrarily long public traces.
Future evaluations can combine \sage{} with text and context compression methods to expose agents to longer histories while keeping inference cost manageable \citep{tang2025perceptioncompressortrainingfreeprompt,tang2026gmsaenhancingcontextcompression,tang2026comicoarsetofinecontextcompression,tang2026readhumancompressingcontext,tang2026positionbiasshiftingcontext}.

\section{Potential Risks}

\sage{} studies public experience sharing in sandboxed benchmark environments, but similar mechanisms could expose sensitive trajectories, prompts, tool outputs, or user data in real deployments.
They may also encourage agents to imitate high-scoring but brittle or unsafe behaviors from peer histories.
Practical uses need to sanitize shared logs, restrict access to sensitive artifacts, and audit whether reused strategies remain safe outside the benchmark setting.

\section{Reproducibility}

We provide the code needed to reproduce all experiments, including arena runners, agent configurations, evaluation scripts, and analysis utilities.
The code is released under the MIT License.

\section{Information About Use of AI Assistants}

AI assistants were used only for language polishing.
They were not used to generate scientific claims, experimental results, analyses, or conclusions.

\clearpage
\bibliography{custom}

\begin{thebibliography}{57}
\providecommand{\natexlab}[1]{#1}

\bibitem[{Abdulhai et~al.(2026)Abdulhai, Cheng, Shrivastava, Kumar, and
  Levine}]{abdulhai2026har}
Marwa Abdulhai, Ryan Cheng, Aryansh Shrivastava, Aviral Kumar, and Sergey
  Levine. 2026.
\newblock \href {https://openreview.net/forum?id=p144zx4bO0} {Hierarchical
  agenda reasoning for strategic multi-turn dialogue agents}.
\newblock In \emph{The 1st Workshop on Scaling Post-training for LLMs}.

\bibitem[{Baker et~al.(2020)Baker, Kanitscheider, Markov, Wu, Powell, McGrew,
  and Mordatch}]{baker2019emergent}
Bowen Baker, Ingmar Kanitscheider, Todor Markov, Yi~Wu, Glenn Powell, Bob
  McGrew, and Igor Mordatch. 2020.
\newblock \href {https://openreview.net/forum?id=SkxpxJBKwS} {Emergent tool use
  from multi-agent autocurricula}.
\newblock In \emph{International Conference on Learning Representations}.

\bibitem[{Bakhtin et~al.(2022)Bakhtin, Brown, Dinan, Farina, Flaherty, Fried,
  Goff, Gray, Hu, Jacob, Komeili, Konath, Kwon, Lerer, Lewis, Miller, Mitts,
  Renduchintala, Roller, Rowe, Shi, Spisak, Wei, Wu, Zhang, and
  Zijlstra}]{brown2022cicero}
Anton Bakhtin, Noam Brown, Emily Dinan, Gabriele Farina, Colin Flaherty, Daniel
  Fried, Andrew Goff, Jonathan Gray, Hengyuan Hu, Athul~Paul Jacob, Mojtaba
  Komeili, Karthik Konath, Minae Kwon, Adam Lerer, Mike Lewis, Alexander~H.
  Miller, Sasha Mitts, Adithya Renduchintala, Stephen Roller, and 7 others.
  2022.
\newblock \href {https://doi.org/10.1126/science.ade9097} {Human-level play in
  the game of {Diplomacy} by combining language models with strategic
  reasoning}.
\newblock \emph{Science}, 378(6624):1067--1074.

\bibitem[{Belle et~al.(2025)Belle, Barnes, Amayuelas, Bercovich, Wang, and
  Wang}]{belle2025agentsofchange}
Nikolas Belle, Dakota Barnes, Alfonso Amayuelas, Ivan Bercovich, Xin~Eric Wang,
  and William Wang. 2025.
\newblock \href {https://arxiv.org/abs/2506.04651} {Agents of change:
  Self-evolving {LLM} agents for strategic planning}.
\newblock \emph{Preprint}, arXiv:2506.04651.

\bibitem[{Blum and Hardt(2015)}]{blum2015ladder}
Avrim Blum and Moritz Hardt. 2015.
\newblock \href {https://proceedings.mlr.press/v37/blum15.html} {The ladder: A
  reliable leaderboard for machine learning competitions}.
\newblock In \emph{Proceedings of the 32nd International Conference on Machine
  Learning}, volume~37 of \emph{Proceedings of Machine Learning Research},
  pages 1006--1014, Lille, France. PMLR.

\bibitem[{Bridgland(2020)}]{m4cs2020drugwars}
Max Bridgland. 2020.
\newblock Drugwars: The {DOS} game from the 80s re-written in python from
  scratch.
\newblock \url{https://github.com/M4cs/drugwars}.
\newblock Open-source software.

\bibitem[{Chan et~al.(2024)Chan, Jiayang, Yim, Deng, Fan, Li, Liu, Zhang, Wang,
  and Song}]{chan2024negotiationtom}
Chunkit Chan, Cheng Jiayang, Yauwai Yim, Zheye Deng, Wei Fan, Haoran Li, Xin
  Liu, Hongming Zhang, Weiqi Wang, and Yangqiu Song. 2024.
\newblock \href {https://doi.org/10.18653/v1/2024.findings-emnlp.244}
  {{N}egotiation{T}o{M}: A benchmark for stress-testing machine theory of mind
  on negotiation surrounding}.
\newblock In \emph{Findings of the Association for Computational Linguistics:
  EMNLP 2024}, pages 4211--4241, Miami, Florida, USA. Association for
  Computational Linguistics.

\bibitem[{Chan et~al.(2025)Chan, Chowdhury, Jaffe, Aung, Sherburn, Mays,
  Starace, Liu, Maksin, Patwardhan, Madry, and Weng}]{chan2025mlebench}
Jun~Shern Chan, Neil Chowdhury, Oliver Jaffe, James Aung, Dane Sherburn, Evan
  Mays, Giulio Starace, Kevin Liu, Leon Maksin, Tejal Patwardhan, Aleksander
  Madry, and Lilian Weng. 2025.
\newblock \href {https://openreview.net/forum?id=6s5uXNWGIh} {{MLE}-bench:
  Evaluating machine learning agents on machine learning engineering}.
\newblock In \emph{The Thirteenth International Conference on Learning
  Representations}.

\bibitem[{Chen et~al.(2025)Chen, Xiong, Lu, Han, Deng, He, Wu, Li, Liu, and
  Hooi}]{chen2025mlrbench}
Hui Chen, Miao Xiong, Yujie Lu, Wei Han, Ailin Deng, Yufei He, Jiaying Wu, Yibo
  Li, Yue Liu, and Bryan Hooi. 2025.
\newblock \href {https://arxiv.org/abs/2505.19955} {{MLR}-bench: Evaluating ai
  agents on open-ended machine learning research}.
\newblock \emph{Preprint}, arXiv:2505.19955.

\bibitem[{Cowboys(2026)}]{spacecowboys2026splendor}
Space Cowboys. 2026.
\newblock Splendor.
\newblock \url{https://www.spacecowboys-games.com/game/splendor/}.
\newblock Official game description.

\bibitem[{DeepSeek-AI et~al.(2025)DeepSeek-AI, Liu, Mei, Lin, Xue, Wang, Xu,
  Wu, Zhang, Lin, Dong, Lu, Zhao, Deng, Xu, Ruan, Dai, Guo, Yang, Chen, Li,
  Zhou, Lin, Dai, Hao, Chen, Li, Zhang, Xu, Li, Liang, Wei, Zhang, Luo, Ji,
  Ding, Tang, Cao, Gao, Qu, Zeng, Huang, Li, Xu, Hu, Chen, Xiang, Yuan, Cheng,
  Zhu, Ran, Jiang, Qiu, Li, Song, Dong, Gao, Guan, Huang, Zhou, Huang, Yu,
  Wang, Zhang, Wang, Zhao, Yin, Guo, Luo, Ma, Wang, Zhang, Di, Xu, Zhang,
  Zhang, Tang, Zhou, Huang, Cong, Wang, Wang, Zhu, Li, Chen, Du, Xu, Ge, Zhang,
  Pan, Wang, Yin, Xu, Shen, Zhang, Liu, Lu, Zhou, Chen, Cai, Chen, Hu, Liu, Hu,
  Ma, Wang, Yu, Zhou, Pan, Zhou, Ni, Yun, Pei, Ye, Yue, Zeng, Liu, Liang, Pang,
  Luo, Gao, Zhang, Gao, Wang, Bi, Liu, Wang, Chen, Zhang, Nie, Cheng, Liu, Xie,
  Liu, Yu, Li, Yang, Li, Chen, Su, Pan, Lin, Fu, Wang, Zhang, Xu, Ma, Li, Li,
  Zhao, Sun, Wang, Qian, Yu, Zhang, Ding, Shi, Xiong, He, Zhou, Zhong, Piao,
  Wang, Chen, Tan, Wei, Ma, Liu, Yang, Guo, Wu, Wu, Cheng, Ou, Xu, Wang, Gong,
  Wu, Zou, Li, Xiong, Luo, You, Liu, Zhou, Wu, Ren, Zhao, Ren, Sha, Fu, Xu,
  Xie, Zhang, Hao, Gou, Ma, Yan, Shao, Huang, Wu, Li, Zhang, Xu, Wang, Gu, Zhu,
  Li, Zhang, Xie, Gao, Pan, Yao, Feng, Li, Cai, Ni, Xu, Li, Tian, Chen, Jin,
  Li, Zhou, Sun, Li, Jin, Shen, Chen, Song, Zhou, Zhu, Huang, Li, Zheng, Zhu,
  Ma, Huang, Xu, Zhang, Ji, Liang, Guo, Chen, Xia, Wang, Li, Zhang, Chen, Sun,
  Wu, Ye, Wang, Xiao, An, Wang, Sun, Wang, Tang, Zha, Zhang, Ju, Zhang, and
  Qu}]{deepseekai2025deepseekv32}
DeepSeek-AI, Aixin Liu, Aoxue Mei, Bangcai Lin, Bing Xue, Bingxuan Wang,
  Bingzheng Xu, Bochao Wu, Bowei Zhang, Chaofan Lin, Chen Dong, Chengda Lu,
  Chenggang Zhao, Chengqi Deng, Chenhao Xu, Chong Ruan, Damai Dai, Daya Guo,
  Dejian Yang, and 245 others. 2025.
\newblock \href {https://arxiv.org/abs/2512.02556} {{DeepSeek-V3.2}: Pushing
  the frontier of open large language models}.
\newblock \emph{Preprint}, arXiv:2512.02556.

\bibitem[{Du et~al.(2024)Du, Li, Torralba, Tenenbaum, and
  Mordatch}]{du2024multiagent}
Yilun Du, Shuang Li, Antonio Torralba, Joshua~B. Tenenbaum, and Igor Mordatch.
  2024.
\newblock \href {https://proceedings.mlr.press/v235/du24e.html} {Improving
  factuality and reasoning in language models through multiagent debate}.
\newblock In \emph{Proceedings of the 41st International Conference on Machine
  Learning}, volume 235 of \emph{Proceedings of Machine Learning Research},
  pages 11733--11763. PMLR.

\bibitem[{Duan et~al.(2024)Duan, Zhang, Diffenderfer, Kailkhura, Sun,
  Stengel-Eskin, Bansal, Chen, and Xu}]{li2024gtbench}
Jinhao Duan, Renming Zhang, James Diffenderfer, Bhavya Kailkhura, Lichao Sun,
  Elias Stengel-Eskin, Mohit Bansal, Tianlong Chen, and Kaidi Xu. 2024.
\newblock \href {https://arxiv.org/abs/2402.12348} {{GTBench}: Uncovering the
  strategic reasoning limitations of llms via game-theoretic evaluations}.
\newblock \emph{Preprint}, arXiv:2402.12348.

\bibitem[{Fu et~al.(2026)Fu, Ding, Pan, Zhu, Zhang, Qiu, Liu, Zhang, Cao, Cai,
  Ding, and Yu}]{fu2025catarena}
Lingyue Fu, Xin Ding, Linyue Pan, Yaoming Zhu, Shao Zhang, Lin Qiu, Weiwen Liu,
  Weinan Zhang, Xuezhi Cao, Xunliang Cai, Jiaxin Ding, and Yong Yu. 2026.
\newblock \href {https://arxiv.org/abs/2510.26852} {{CATArena}: Evaluating
  evolutionary capabilities of code agents via iterative tournaments}.
\newblock \emph{Preprint}, arXiv:2510.26852.

\bibitem[{Google(2025)}]{googledeepmind2025gemini3flash}
Google. 2025.
\newblock {Gemini 3 Flash Preview}.
\newblock
  \url{https://ai.google.dev/gemini-api/docs/models/gemini-3-flash-preview}.
\newblock Gemini API model documentation.

\bibitem[{Google(2026)}]{google2026adk}
Google. 2026.
\newblock Agent development kit.
\newblock \url{https://adk.dev/}.
\newblock Developer documentation.

\bibitem[{Huang et~al.(2024)Huang, Vora, Liang, and
  Leskovec}]{huang2024mlagentbench}
Qian Huang, Jian Vora, Percy Liang, and Jure Leskovec. 2024.
\newblock \href {https://proceedings.mlr.press/v235/huang24y.html}
  {{MLA}gent{B}ench: Evaluating language agents on machine learning
  experimentation}.
\newblock In \emph{Proceedings of the 41st International Conference on Machine
  Learning}, volume 235 of \emph{Proceedings of Machine Learning Research},
  pages 20271--20309. PMLR.

\bibitem[{Jaderberg et~al.(2017)Jaderberg, Dalibard, Osindero, Czarnecki,
  Donahue, Razavi, Vinyals, Green, Dunning, Simonyan, Fernando, and
  Kavukcuoglu}]{jaderberg2017population}
Max Jaderberg, Valentin Dalibard, Simon Osindero, Wojciech~M. Czarnecki, Jeff
  Donahue, Ali Razavi, Oriol Vinyals, Tim Green, Iain Dunning, Karen Simonyan,
  Chrisantha Fernando, and Koray Kavukcuoglu. 2017.
\newblock \href {https://arxiv.org/abs/1711.09846} {Population based training
  of neural networks}.
\newblock \emph{Preprint}, arXiv:1711.09846.

\bibitem[{Kiela et~al.(2021)Kiela, Bartolo, Nie, Kaushik, Geiger, Wu, Vidgen,
  Prasad, Singh, Ringshia, Ma, Thrush, Riedel, Waseem, Stenetorp, Jia, Bansal,
  Potts, and Williams}]{kiela2021dynabench}
Douwe Kiela, Max Bartolo, Yixin Nie, Divyansh Kaushik, Atticus Geiger,
  Zhengxuan Wu, Bertie Vidgen, Grusha Prasad, Amanpreet Singh, Pratik Ringshia,
  Zhiyi Ma, Tristan Thrush, Sebastian Riedel, Zeerak Waseem, Pontus Stenetorp,
  Robin Jia, Mohit Bansal, Christopher Potts, and Adina Williams. 2021.
\newblock \href {https://doi.org/10.18653/v1/2021.naacl-main.324} {Dynabench:
  Rethinking benchmarking in {NLP}}.
\newblock In \emph{Proceedings of the 2021 Conference of the North American
  Chapter of the Association for Computational Linguistics: Human Language
  Technologies}, pages 4110--4124, Online. Association for Computational
  Linguistics.

\bibitem[{Li et~al.(2023)Li, Hammoud, Itani, Khizbullin, and
  Ghanem}]{li2023camel}
Guohao Li, Hasan Abed Al~Kader Hammoud, Hani Itani, Dmitrii Khizbullin, and
  Bernard Ghanem. 2023.
\newblock \href {https://arxiv.org/abs/2303.17760} {Camel: Communicative agents
  for "mind" exploration of large language model society}.
\newblock \emph{Preprint}, arXiv:2303.17760.

\bibitem[{Li et~al.(2025)Li, Lin, Xia, and Jin}]{li2025rethinkingmoa}
Wenzhe Li, Yong Lin, Mengzhou Xia, and Chi Jin. 2025.
\newblock \href {https://arxiv.org/abs/2502.00674} {Rethinking
  mixture-of-agents: Is mixing different large language models beneficial?}
\newblock \emph{Preprint}, arXiv:2502.00674.

\bibitem[{Light et~al.(2023)Light, Cai, Shen, and Hu}]{cheng2023avalonbench}
Jonathan Light, Min Cai, Sheng Shen, and Ziniu Hu. 2023.
\newblock \href {https://arxiv.org/abs/2310.05036} {Avalonbench: Evaluating
  llms playing the game of avalon}.
\newblock \emph{Preprint}, arXiv:2310.05036.

\bibitem[{Lupu et~al.(2025)Lupu, Willi, and Foerster}]{lupu2025decrypto}
Andrei Lupu, Timon Willi, and Jakob Foerster. 2025.
\newblock \href {https://arxiv.org/abs/2506.20664} {The decrypto benchmark for
  multi-agent reasoning and theory of mind}.
\newblock \emph{Preprint}, arXiv:2506.20664.

\bibitem[{Madaan et~al.(2023)Madaan, Tandon, Gupta, Hallinan, Gao, Wiegreffe,
  Alon, Dziri, Prabhumoye, Yang, Gupta, Majumder, Hermann, Welleck,
  Yazdanbakhsh, and Clark}]{madaan2023selfrefine}
Aman Madaan, Niket Tandon, Prakhar Gupta, Skyler Hallinan, Luyu Gao, Sarah
  Wiegreffe, Uri Alon, Nouha Dziri, Shrimai Prabhumoye, Yiming Yang, Shashank
  Gupta, Bodhisattwa~Prasad Majumder, Katherine Hermann, Sean Welleck, Amir
  Yazdanbakhsh, and Peter Clark. 2023.
\newblock \href {https://openreview.net/forum?id=S37hOerQLB} {Self-refine:
  Iterative refinement with self-feedback}.
\newblock In \emph{Thirty-seventh Conference on Neural Information Processing
  Systems}.

\bibitem[{Maharana et~al.(2024)Maharana, Lee, Tulyakov, Bansal, Barbieri, and
  Fang}]{maharana2024locomo}
Adyasha Maharana, Dong-Ho Lee, Sergey Tulyakov, Mohit Bansal, Francesco
  Barbieri, and Yuwei Fang. 2024.
\newblock \href {https://doi.org/10.18653/v1/2024.acl-long.747} {Evaluating
  very long-term conversational memory of {LLM} agents}.
\newblock In \emph{Proceedings of the 62nd Annual Meeting of the Association
  for Computational Linguistics (Volume 1: Long Papers)}, pages 13851--13870,
  Bangkok, Thailand. Association for Computational Linguistics.

\bibitem[{Nathani et~al.(2025)Nathani, Madaan, Roberts, Bashlykov, Menon,
  Moens, Budhiraja, Magka, Vorotilov, Chaurasia, Hupkes, Cabral, Shavrina,
  Foerster, Bachrach, Wang, and Raileanu}]{nathani2025mlgym}
Deepak Nathani, Lovish Madaan, Nicholas Roberts, Nikolay Bashlykov, Ajay Menon,
  Vincent Moens, Amar Budhiraja, Despoina Magka, Vladislav Vorotilov, Gaurav
  Chaurasia, Dieuwke Hupkes, Ricardo~Silveira Cabral, Tatiana Shavrina, Jakob
  Foerster, Yoram Bachrach, William~Yang Wang, and Roberta Raileanu. 2025.
\newblock \href {https://arxiv.org/abs/2502.14499} {Mlgym: A new framework and
  benchmark for advancing ai research agents}.
\newblock \emph{Preprint}, arXiv:2502.14499.

\bibitem[{Ndousse et~al.(2021)Ndousse, Eck, Levine, and
  Jaques}]{ndousse2021emergent}
Kamal~K Ndousse, Douglas Eck, Sergey Levine, and Natasha Jaques. 2021.
\newblock \href {https://proceedings.mlr.press/v139/ndousse21a.html} {Emergent
  social learning via multi-agent reinforcement learning}.
\newblock In \emph{Proceedings of the 38th International Conference on Machine
  Learning}, volume 139 of \emph{Proceedings of Machine Learning Research},
  pages 7991--8004. PMLR.

\bibitem[{Novikov et~al.(2025)Novikov, V{\~u}, Eisenberger, Dupont, Huang,
  Wagner, Shirobokov, Kozlovskii, Ruiz, Mehrabian, Kumar, See, Chaudhuri,
  Holland, Davies, Nowozin, Kohli, and Balog}]{novikov2025alphaevolve}
Alexander Novikov, Ng{\^a}n V{\~u}, Marvin Eisenberger, Emilien Dupont, Po-Sen
  Huang, Adam~Zsolt Wagner, Sergey Shirobokov, Borislav Kozlovskii, Francisco
  J.~R. Ruiz, Abbas Mehrabian, M.~Pawan Kumar, Abigail See, Swarat Chaudhuri,
  George Holland, Alex Davies, Sebastian Nowozin, Pushmeet Kohli, and Matej
  Balog. 2025.
\newblock \href {https://arxiv.org/abs/2506.13131} {Alphaevolve: A coding agent
  for scientific and algorithmic discovery}.
\newblock \emph{Preprint}, arXiv:2506.13131.

\bibitem[{OpenAI(2026)}]{openai2026gpt54}
OpenAI. 2026.
\newblock Introducing {GPT-5.4}.
\newblock \url{https://openai.com/index/introducing-gpt-5-4/}.
\newblock Official release note.

\bibitem[{Park et~al.(2023)Park, O'Brien, Cai, Morris, Liang, and
  Bernstein}]{park2023generative}
Joon~Sung Park, Joseph O'Brien, Carrie~Jun Cai, Meredith~Ringel Morris, Percy
  Liang, and Michael~S. Bernstein. 2023.
\newblock \href {https://doi.org/10.1145/3586183.3606763} {Generative agents:
  Interactive simulacra of human behavior}.
\newblock In \emph{Proceedings of the 36th Annual ACM Symposium on User
  Interface Software and Technology}, UIST '23, pages 1--22. ACM.

\bibitem[{Reloaded(2026)}]{drugwars2026reloaded}
Drug~Wars: Reloaded. 2026.
\newblock Drug wars: Reloaded.
\newblock \url{https://drugwars.app/}.
\newblock Game description.

\bibitem[{Robeyns et~al.(2025)Robeyns, Szummer, and
  Aitchison}]{robeyns2025selfimproving}
Maxime Robeyns, Martin Szummer, and Laurence Aitchison. 2025.
\newblock \href {https://openreview.net/forum?id=rShJCyLsOr} {A self-improving
  coding agent}.
\newblock In \emph{Scaling Self-Improving Foundation Models without Human
  Supervision}.

\bibitem[{Seed(2026)}]{bytedanceseed2026seed20}
ByteDance Seed. 2026.
\newblock {Seed2.0}.
\newblock \url{https://research.doubao.com/en/seed2}.
\newblock Model card and technical blog.

\bibitem[{Shinn et~al.(2023)Shinn, Cassano, Berman, Gopinath, Narasimhan, and
  Yao}]{shinn2023reflexion}
Noah Shinn, Federico Cassano, Edward Berman, Ashwin Gopinath, Karthik
  Narasimhan, and Shunyu Yao. 2023.
\newblock \href {https://arxiv.org/abs/2303.11366} {Reflexion: Language agents
  with verbal reinforcement learning}.
\newblock \emph{Preprint}, arXiv:2303.11366.

\bibitem[{Smith et~al.(2026)Smith, Abdulhai, Diaz, Tesic, Trivedi, Vezhnevets,
  Hammond, Clifton, Chang, Du{\'e}{\~n}ez-Guzm{\'a}n, Agapiou, Matyas, Karmon,
  Zhang, Dilkes, Kundu, Nguyen, Tewolde, Purbey, Kadiyala, Gupta, Korshuk,
  Alexander, Makarov, Zhao, Fernandez, Wang, Wang, Cui, Xiao, Shi, Sung,
  Rahman, Stone, Kang, Yun, Ananya, Cha, Wu, Tennant, Macmillan-Scott, Segura,
  Riazi, Cui, Subramanian, Klassen, Schiavone, Alim, McIlraith, Beltran,
  Pe{\~n}a, Rojas, Chacon-Chamorro, Manrique, Giraldo, Quijano, Wang, Chen,
  Zhong, Wang, Tu, Zhang, Chen, Jia, Feng, Zheng, Lin, Fan, Liu, Sarangi, Wang,
  Shi, Du, Kulandaivel, Liu, Ruiyang, Talele, Lu, Piqueras, Dhuri, McHale,
  Baarslag, Hadfield-Menell, Jaques, Hernandez-Orallo, and
  Leibo}]{smith2025concordia}
Chandler Smith, Marwa Abdulhai, Manfred Diaz, Marko Tesic, Rakshit Trivedi,
  Sasha Vezhnevets, Lewis Hammond, Jesse Clifton, Minsuk Chang, Edgar~A.
  Du{\'e}{\~n}ez-Guzm{\'a}n, John~P Agapiou, Jayd Matyas, Danny Karmon, Beining
  Zhang, Jim Dilkes, Akash Kundu, Jord Nguyen, Emanuel Tewolde, Jebish Purbey,
  and 67 others. 2026.
\newblock \href {https://openreview.net/forum?id=yG4Fj0voJZ} {Evaluating
  generalization capabilities of {LLM}-based agents in mixed-motive scenarios
  using concordia}.
\newblock In \emph{The Thirty-ninth Annual Conference on Neural Information
  Processing Systems Datasets and Benchmarks Track}.

\bibitem[{Tan et~al.(2025)Tan, Yan, Hsu, Han, Wang, Le, Song, Chen, Palangi,
  Lee, Iyer, Chen, Liu, Lee, and Pfister}]{tan2025reflectivememory}
Zhen Tan, Jun Yan, I-Hung Hsu, Rujun Han, Zifeng Wang, Long Le, Yiwen Song,
  Yanfei Chen, Hamid Palangi, George Lee, Anand~Rajan Iyer, Tianlong Chen, Huan
  Liu, Chen-Yu Lee, and Tomas Pfister. 2025.
\newblock \href {https://doi.org/10.18653/v1/2025.acl-long.413} {In prospect
  and retrospect: Reflective memory management for long-term personalized
  dialogue agents}.
\newblock In \emph{Proceedings of the 63rd Annual Meeting of the Association
  for Computational Linguistics (Volume 1: Long Papers)}, pages 8416--8439,
  Vienna, Austria. Association for Computational Linguistics.

\bibitem[{Tang et~al.(2026{\natexlab{a}})Tang, Huang, Zhang, Zhang, Yu, Zheng,
  Meng, and Yin}]{tang2026positionbiasshiftingcontext}
Jiwei Tang, Zhijing Huang, Xinyu Zhang, Chen~Jason Zhang, Jianxing Yu, Libin
  Zheng, Rui Meng, and Jian Yin. 2026{\natexlab{a}}.
\newblock \href {https://arxiv.org/abs/2605.09463} {Beyond position bias:
  Shifting context compression from position-driven to semantic-driven}.
\newblock \emph{Preprint}, arXiv:2605.09463.

\bibitem[{Tang et~al.(2026{\natexlab{b}})Tang, Liu, Zhang, Lv, Zhao, Lu, Liu,
  Chen, Yuan, Zheng, Su, and Zheng}]{tang2026readhumancompressingcontext}
Jiwei Tang, Shilei Liu, Zhicheng Zhang, Qingsong Lv, Runsong Zhao, Tingwei Lu,
  Langming Liu, Haibin Chen, Yujin Yuan, Hai-Tao Zheng, Wenbo Su, and Bo~Zheng.
  2026{\natexlab{b}}.
\newblock \href {https://arxiv.org/abs/2602.01840} {Read as human: Compressing
  context via parallelizable close reading and skimming}.
\newblock \emph{Preprint}, arXiv:2602.01840.

\bibitem[{Tang et~al.(2026{\natexlab{c}})Tang, Liu, Zhang, Yuan, Zheng, Su, and
  Zheng}]{tang2026comicoarsetofinecontextcompression}
Jiwei Tang, Shilei Liu, Zhicheng Zhang, Yujin Yuan, Libin Zheng, Wenbo Su, and
  Bo~Zheng. 2026{\natexlab{c}}.
\newblock \href {https://arxiv.org/abs/2602.01719} {Comi: Coarse-to-fine
  context compression via marginal information gain}.
\newblock \emph{Preprint}, arXiv:2602.01719.

\bibitem[{Tang et~al.(2025)Tang, Xu, Lu, Zhang, Zhao, Hai, and
  Zheng}]{tang2025perceptioncompressortrainingfreeprompt}
Jiwei Tang, Jin Xu, Tingwei Lu, Zhicheng Zhang, Yiming Zhao, Lin Hai, and
  Hai-Tao Zheng. 2025.
\newblock \href {https://arxiv.org/abs/2409.19272} {Perception compressor: A
  training-free prompt compression framework in long context scenarios}.
\newblock \emph{Preprint}, arXiv:2409.19272.

\bibitem[{Tang et~al.(2026{\natexlab{d}})Tang, Zhang, Wu, Ye, Bai, Wang, Lu,
  Hai, Zhao, Zheng, and Kim}]{tang2026gmsaenhancingcontextcompression}
Jiwei Tang, Zhicheng Zhang, Shunlong Wu, Jingheng Ye, Lichen Bai, Zitai Wang,
  Tingwei Lu, Lin Hai, Yiming Zhao, Hai-Tao Zheng, and Hong-Gee Kim.
  2026{\natexlab{d}}.
\newblock \href {https://arxiv.org/abs/2505.12215} {Gmsa: Enhancing context
  compression via group merging and layer semantic alignment}.
\newblock \emph{Preprint}, arXiv:2505.12215.

\bibitem[{Team et~al.(2026)Team, Bai, Bai, Bao, Cai, Cao, Charles, Che, Chen,
  Chen, Chen, Chen, Chen, Chen, Chen, Chen, Chen, Chen, Chen, Chen, Chen, Chen,
  Chen, Chen, Chen, Chen, Chen, Chen, Chen, Cheng, Chu, Cui, Deng, Diao, Ding,
  Dong, Dong, Dong, Dong, Du, Du, Du, Du, Du, Fan, Fang, Feng, Feng, Fu, Fu,
  Gao, Gao, Ge, Geng, Gong, Gong, Gongque, Gu, Gu, Gu, Guan, Guo, Hao, He, He,
  He, Hong, Hu, Hu, Hu, Hu, Huang, Huang, Huang, Huang, Jiang, Jiang, Jin,
  Jing, Lai, Li, Li, Li, Li, Li, Li, Li, Li, Li, Li, Li, Li, Li, Li, Li, Li,
  Li, Li, Li, Li, Li, Li, Li, Liao, Lin, Lin, Lin, Lin, Liu, Liu, Liu, Liu,
  Liu, Liu, Liu, Liu, Liu, Liu, Liu, Liu, Liu, Liu, Liu, Liu, Liu, Liu, Lu, Lu,
  Lu, Luo, Luo, Luo, Ma, Ma, Mao, Mei, Men, Meng, Meng, Miao, Ni, Ouyang, Pan,
  Pang, Qian, Qin, Qin, Qiu, Qu, Shang, Shao, Shen, Shen, Shi, Shi, Shi, Song,
  Song, Song, Song, Su, Su, Su, Sui, Sun, Sun, Sun, Sung, Tai, Tang, Tang,
  Tang, Tang, Tao, Teng, Tian, Tian, Wang, Wang, Wang, Wang, Wang, Wang, Wang,
  Wang, Wang, Wang, Wang, Wang, Wang, Wang, Wang, Wang, Wang, Wang, Wang, Wang,
  Wang, Wang, Wang, Wang, Wang, Wang, Wang, Wang, Wang, Wang, Wang, Wang, Wang,
  Wang, Wang, Wang, Wang, Wang, Wei, Wei, Wen, Wen, Wu, Wu, Wu, Wu, Wu, Wu, Wu,
  Wu, Wu, Xiao, Xie, Xie, Xie, Xin, Xing, Xu, Xu, Xu, Xu, Xu, Xu, Xu, Xu, Xu,
  Xu, Xu, Xu, Xu, Xu, Xu, Yan, Yan, Yang, Yang, Yang, Yang, Yang, Yang, Yang,
  Yang, Yang, Yang, Yang, Yang, Yang, Yang, Yao, Ye, Ye, Ye, Yin, Yu, Yu, Yu,
  Yu, Yuan, Yuan, Yuan, Yue, Zeng, Zha, Zhan, Zhang, Zhang, Zhang, Zhang,
  Zhang, Zhang, Zhang, Zhang, Zhang, Zhang, Zhang, Zhang, Zhang, Zhang, Zhang,
  Zhang, Zhang, Zhang, Zhao, Zhao, Zhao, Zhao, Zhao, Zhao, Zhao, Zheng, Zheng,
  Zheng, Zheng, Zhong, Zhong, Zhong, Zhou, Zhou, Zhou, Zhou, Zhu, Zhu, Zhu,
  Zhu, Zhu, Zhuang, Zhuang, Zou, and Zu}]{moonshotai2026kimi}
Kimi Team, Tongtong Bai, Yifan Bai, Yiping Bao, S.~H. Cai, Yuan Cao,
  Y.~Charles, H.~S. Che, Cheng Chen, Guanduo Chen, Huarong Chen, Jia Chen,
  Jiahao Chen, Jianlong Chen, Jun Chen, Kefan Chen, Liang Chen, Ruijue Chen,
  Xinhao Chen, and 307 others. 2026.
\newblock \href {https://arxiv.org/abs/2602.02276} {{Kimi K2.5}: Visual agentic
  intelligence}.
\newblock \emph{Preprint}, arXiv:2602.02276.

\bibitem[{Vinyals et~al.(2019)Vinyals, Babuschkin, Czarnecki, Mathieu, Dudzik,
  Chung, Choi, Powell, Ewalds, Georgiev, Oh, Horgan, Kroiss, Danihelka, Huang,
  Sifre, Cai, Agapiou, Jaderberg, Vezhnevets, Leblond, Pohlen, Dalibard,
  Budden, Sulsky, Molloy, Paine, Gulcehre, Wang, Pfaff, Wu, Ring, Yogatama,
  W{\"u}nsch, McKinney, Smith, Schaul, Lillicrap, Kavukcuoglu, Hassabis, Apps,
  and Silver}]{vinyals2019grandmaster}
Oriol Vinyals, Igor Babuschkin, Wojciech~M. Czarnecki, Micha{\"e}l Mathieu,
  Andrew Dudzik, Junyoung Chung, David~H. Choi, Richard Powell, Timo Ewalds,
  Petko Georgiev, Junhyuk Oh, Dan Horgan, Manuel Kroiss, Ivo Danihelka, Aja
  Huang, Laurent Sifre, Trevor Cai, John~P. Agapiou, Max Jaderberg, and 23
  others. 2019.
\newblock \href {https://doi.org/10.1038/s41586-019-1724-z} {Grandmaster level
  in {StarCraft II} using multi-agent reinforcement learning}.
\newblock \emph{Nature}, 575(7782):350--354.

\bibitem[{Wang et~al.(2023)Wang, Xie, Jiang, Mandlekar, Xiao, Zhu, Fan, and
  Anandkumar}]{wang2023voyager}
Guanzhi Wang, Yuqi Xie, Yunfan Jiang, Ajay Mandlekar, Chaowei Xiao, Yuke Zhu,
  Linxi Fan, and Anima Anandkumar. 2023.
\newblock \href {https://arxiv.org/abs/2305.16291} {Voyager: An open-ended
  embodied agent with large language models}.
\newblock \emph{Preprint}, arXiv:2305.16291.

\bibitem[{Wang et~al.(2024)Wang, Wang, Su, Tong, and Song}]{wang2024rethinking}
Qineng Wang, Zihao Wang, Ying Su, Hanghang Tong, and Yangqiu Song. 2024.
\newblock \href {https://arxiv.org/abs/2402.18272} {Rethinking the bounds of
  {LLM} reasoning: Are multi-agent discussions the key?}
\newblock \emph{Preprint}, arXiv:2402.18272.

\bibitem[{Wei et~al.(2026)Wei, Sachdeva, Coleman, He, Bei, Ning, Ai, Li, He,
  Chi, Wang, Chen, Pereira, Kang, and Cheng}]{wei2025evomemory}
Tianxin Wei, Noveen Sachdeva, Benjamin Coleman, Zhankui He, Yuanchen Bei,
  Xuying Ning, Mengting Ai, Yunzhe Li, Jingrui He, Ed~H. Chi, Chi Wang, Shuo
  Chen, Fernando Pereira, Wang-Cheng Kang, and Derek~Zhiyuan Cheng. 2026.
\newblock \href {https://arxiv.org/abs/2511.20857} {Evo-memory: Benchmarking
  {LLM} agent test-time learning with self-evolving memory}.
\newblock \emph{Preprint}, arXiv:2511.20857.

\bibitem[{Weng et~al.(2026)Weng, Antoniades, Nathani, Zhang, Pu, and
  Wang}]{weng2026group}
Zhaotian Weng, Antonis Antoniades, Deepak Nathani, Zhen Zhang, Xiao Pu, and
  Xin~Eric Wang. 2026.
\newblock \href {https://arxiv.org/abs/2602.04837} {Group-evolving agents:
  Open-ended self-improvement via experience sharing}.
\newblock \emph{Preprint}, arXiv:2602.04837.

\bibitem[{Wijk et~al.(2025)Wijk, Lin, Becker, Jawhar, Parikh, Broadley, Chan,
  Chen, Clymer, Dhyani, Ericheva, Garcia, Goodrich, Jurkovic, Kinniment, Lajko,
  Nix, Sato, Saunders, Taran, West, and Barnes}]{wijk2025rebench}
Hjalmar Wijk, Tao~Roa Lin, Joel Becker, Sami Jawhar, Neev Parikh, Thomas
  Broadley, Lawrence Chan, Michael Chen, Joshua~M Clymer, Jai Dhyani, Elena
  Ericheva, Katharyn Garcia, Brian Goodrich, Nikola Jurkovic, Megan Kinniment,
  Aron Lajko, Seraphina Nix, Lucas Jun~Koba Sato, William Saunders, and 3
  others. 2025.
\newblock \href {https://openreview.net/forum?id=3rB0bVU6z6} {{RE}-bench:
  Evaluating frontier {AI} r\&d capabilities of language model agents against
  human experts}.
\newblock In \emph{Forty-second International Conference on Machine Learning}.

\bibitem[{Wu et~al.(2025)Wu, Wang, Yu, Zhang, Chang, and
  Yu}]{wu2024longmemeval}
Di~Wu, Hongwei Wang, Wenhao Yu, Yuwei Zhang, Kai-Wei Chang, and Dong Yu. 2025.
\newblock \href {https://arxiv.org/abs/2410.10813} {Longmemeval: Benchmarking
  chat assistants on long-term interactive memory}.
\newblock \emph{Preprint}, arXiv:2410.10813.

\bibitem[{Wu et~al.(2023)Wu, Bansal, Zhang, Wu, Li, Zhu, Jiang, Zhang, Zhang,
  Liu, Awadallah, White, Burger, and Wang}]{wu2023autogen}
Qingyun Wu, Gagan Bansal, Jieyu Zhang, Yiran Wu, Beibin Li, Erkang Zhu,
  Li~Jiang, Xiaoyun Zhang, Shaokun Zhang, Jiale Liu, Ahmed~Hassan Awadallah,
  Ryen~W White, Doug Burger, and Chi Wang. 2023.
\newblock \href {https://arxiv.org/abs/2308.08155} {Autogen: Enabling next-gen
  {LLM} applications via multi-agent conversation}.
\newblock \emph{Preprint}, arXiv:2308.08155.

\bibitem[{Xie et~al.(2026)Xie, Guo, Zhang, Xia, Yang, Ma, Yan, and
  Ren}]{xie2026memevobench}
Weiwei Xie, Shaoxiong Guo, Fan Zhang, Tian Xia, Xue Yang, Lizhuang Ma, Junchi
  Yan, and Qibing Ren. 2026.
\newblock \href {https://arxiv.org/abs/2604.15774} {Memevobench: Benchmarking
  safety risks from memory misevolution in {LLM} agents}.
\newblock \emph{Preprint}, arXiv:2604.15774.

\bibitem[{Xiong et~al.(2025)Xiong, Lin, Xie, He, Liu, Tang, Lakkaraju, and
  Xiang}]{xiong2025memorymanagement}
Zidi Xiong, Yuping Lin, Wenya Xie, Pengfei He, Zirui Liu, Jiliang Tang,
  Himabindu Lakkaraju, and Zhen Xiang. 2025.
\newblock \href {https://arxiv.org/abs/2505.16067} {How memory management
  impacts {LLM} agents: An empirical study of experience-following behavior}.
\newblock \emph{Preprint}, arXiv:2505.16067.

\bibitem[{Yuksekgonul et~al.(2026)Yuksekgonul, Koceja, Li, Bianchi, McCaleb,
  Wang, Kautz, Choi, Zou, Guestrin, and Sun}]{yuksekgonul2026learning}
Mert Yuksekgonul, Daniel Koceja, Xinhao Li, Federico Bianchi, Jed McCaleb,
  Xiaolong Wang, Jan Kautz, Yejin Choi, James Zou, Carlos Guestrin, and Yu~Sun.
  2026.
\newblock \href {https://arxiv.org/abs/2601.16175} {Learning to discover at
  test time}.
\newblock \emph{Preprint}, arXiv:2601.16175.

\bibitem[{Zhang et~al.(2026{\natexlab{a}})Zhang, Hu, Lu, Lange, and
  Clune}]{zhang2025darwin}
Jenny Zhang, Shengran Hu, Cong Lu, Robert Lange, and Jeff Clune.
  2026{\natexlab{a}}.
\newblock \href {https://arxiv.org/abs/2505.22954} {Darwin godel machine:
  Open-ended evolution of self-improving agents}.
\newblock \emph{Preprint}, arXiv:2505.22954.

\bibitem[{Zhang et~al.(2026{\natexlab{b}})Zhang, Khalifa, Bhushan, Murphy,
  Logeswaran, Kim, Lee, Lee, and Wang}]{zhang2025mlrcbench}
Yunxiang Zhang, Muhammad Khalifa, Shitanshu Bhushan, Grant~D Murphy, Lajanugen
  Logeswaran, Jaekyeom Kim, Moontae Lee, Honglak Lee, and Lu~Wang.
  2026{\natexlab{b}}.
\newblock \href {https://openreview.net/forum?id=t8Okk2PRWU} {{MLRC}-bench: Can
  language agents solve machine learning research challenges?}
\newblock In \emph{The Thirty-ninth Annual Conference on Neural Information
  Processing Systems Datasets and Benchmarks Track}.

\bibitem[{Zhao et~al.(2024)Zhao, Huang, Xu, Lin, Liu, and
  Huang}]{zhao2023expel}
Andrew Zhao, Daniel Huang, Quentin Xu, Matthieu Lin, Yong-Jin Liu, and Gao
  Huang. 2024.
\newblock \href {https://arxiv.org/abs/2308.10144} {Expel: {LLM} agents are
  experiential learners}.
\newblock \emph{Preprint}, arXiv:2308.10144.

\bibitem[{Zhuge et~al.(2026)Zhuge, Liu, Faccio, Ashley, Csord{\'a}s,
  Gopalakrishnan, Hamdi, Hammoud, Herrmann, Irie, Kirsch, Li, Li, Liu, Mai,
  Pi{\k{e}}kos, Ramesh, Schlag, Shi, Stani{\'c}, Wang, Wang, Xu, Fan, Ghanem,
  and Schmidhuber}]{zhuge2023mindstorms}
Mingchen Zhuge, Haozhe Liu, Francesco Faccio, Dylan~R. Ashley, R{\'o}bert
  Csord{\'a}s, Anand Gopalakrishnan, Abdullah Hamdi, Hasan Abed Al~Kader
  Hammoud, Vincent Herrmann, Kazuki Irie, Louis Kirsch, Bing Li, Guohao Li,
  Shuming Liu, Jinjie Mai, Piotr Pi{\k{e}}kos, Aditya Ramesh, Imanol Schlag,
  Weimin Shi, and 7 others. 2026.
\newblock \href {https://doi.org/10.26599/CVM.2025.9450460} {Mindstorms in
  natural language-based societies of mind}.
\newblock \emph{Preprint}, arXiv:2305.17066.

\end{thebibliography}

\onecolumn
\appendix

\section{RQ1 Numeric Summary}

\begin{table}[H]
\centering
\scriptsize
\caption{RQ1 summary over post-initial evolution rounds.
\socialevo{} and \selfevo{} cells report means of per-round means.
$\Delta$ is \socialevo{} minus \selfevo{}, with a 95\% confidence interval estimated from paired per-round differences.}
\label{tab:rq1-main}
\setlength{\tabcolsep}{3pt}
\resizebox{\textwidth}{!}{%
\begin{tabular}{lcccc|cccc}
\toprule
& \multicolumn{4}{c|}{MLR-Bench} & \multicolumn{4}{c}{DrugWars} \\
\cmidrule(lr){2-5}\cmidrule(lr){6-9}
Agent & \socialevo{} & \selfevo{} & $\Delta$ & 95\% CI & \socialevo{} & \selfevo{} & $\Delta$ & 95\% CI \\
\midrule
DeepSeek-V3.2 & $5.50$ & $5.31$ & $+0.19$ & $[-0.20, +0.58]$ & $14{,}421.2$ & $-1{,}506.9$ & $+15{,}928.1$ & $[+2{,}887.4, +28{,}968.8]$ \\
Doubao & $4.36$ & $4.43$ & $-0.08$ & $[-0.66, +0.51]$ & $40{,}631.7$ & $28{,}741.6$ & $+11{,}890.1$ & $[-10{,}454.4, +34{,}234.7]$ \\
Gemini & $5.03$ & $5.17$ & $-0.14$ & $[-0.70, +0.42]$ & $57{,}334.4$ & $37{,}618.9$ & $+19{,}715.6$ & $[-16{,}070.6, +55{,}501.7]$ \\
GPT-5.4 & $8.02$ & $7.89$ & $+0.13$ & $[-0.13, +0.39]$ & $37{,}421.7$ & $16{,}448.1$ & $+20{,}973.6$ & $[+3{,}776.5, +38{,}170.7]$ \\
Kimi-K2.5 & $5.24$ & $4.91$ & $+0.34$ & $[-0.20, +0.87]$ & $15{,}185.4$ & $128{,}805.7$ & $-113{,}620.3$ & $[-186{,}774.2, -40{,}466.4]$ \\
\bottomrule
\end{tabular}}
\end{table}

\section{Arena Implementation and Inference Details}

\paragraph{DrugWars.}
DrugWars uses 30 in-game days and normal difficulty.
Each agent starts in the Bronx with \$5{,}000 cash, \$0 bank balance, \$5{,}000 debt, 20 health, no guns, no inventory or stash, and trench-coat capacity 100.
Bank balances earn 3\% interest and debt accrues 4\% interest whenever travel advances the day.
The market has six goods and six locations.
The unmodified base price ranges are \$15{,}000--\$29{,}999 for cocaine, \$5{,}000--\$13{,}999 for heroin, \$1{,}000--\$4{,}999 for acid, \$300--\$899 for weed, \$90--\$249 for speed, and \$10--\$89 for ludes.
The implementation samples one daily global price table from ranges shrunk toward each midpoint by factor 0.70 and copies that table to all six locations.
From day 1 onward, the price-news process shuffles possible anomalies and applies at most one anomaly per day.
The anomaly multipliers are 1.7 for police-driven spikes, 2.3 for demand booms, and 0.60 for cheap-supply events.
Travel advances exactly one day and may trigger one pre-sampled encounter among police chases, gun offers, mugging, found drugs, and coat upgrades.
Mugging immediately removes 10\% of cash, found-drug events immediately add 1--6 units when space permits, and police, gun, and coat events become pending events that block all other tools until \texttt{resolve\_event}.
The primary leaderboard metric is \texttt{cash\_after\_debt} $= \texttt{money} + \texttt{bank} - \texttt{debt}$.
The liquidation value is \texttt{cash\_after\_debt} plus the inventory and stash values at the current day and location prices.
All prices and encounters are pre-sampled from the round seed, all agents in a round share the same world, and round $r$ uses base seed $+\,r-1$.

\paragraph{Splendor.}
Splendor uses five league players and 16 arena rounds.
Each arena round follows a five-game leave-one-out schedule.
In each game, one player is held out and the other four players complete a Splendor match, so every player is held out once and participates in four games per arena round.
This schedule makes every pair of players co-occur in three games per arena round.
We compute pairwise outcomes only from games in which both players appear: ties use average ranks, and a pairwise win is counted only when one player has a strictly better rank than the other.
Each game ends when a player reaches 15 raw points, the match budget expires, or any player reaches the 30-turn cap.

\paragraph{MLR-Bench.}
Due to budget constraints, we randomly sample one MLR-Bench task for the experiments: task id 142, \texttt{neurips2023\_regml}.
For experimental convenience, we adapt the MLR-Bench agent harness by translating the original harness logic into natural-language skill instructions and running agents with the Claude Code framework.
The same sampled task is used for the five \socialevo{} repeats and the five compute-matched \selfevo{} runs.
The judge models are \texttt{gemini-3-flash-preview} and \texttt{kimi-k2.5}.

\paragraph{Inference configuration.}
For DrugWars and Splendor, each model request has a 600-second timeout, and context trimming uses arena-specific thresholds.
The context-trimming thresholds are 96{,}000 tokens for DeepSeek-V3.2, 224{,}000 tokens for Doubao, GPT-5.4, and Kimi-K2.5, and 1{,}016{,}576 tokens for Gemini-3 Flash.
These thresholds reserve output space during context trimming and should not be read as explicit \texttt{max\_tokens} arguments.
ADK calls use up to five attempts with exponential backoff from 0.5 to 10 seconds.
DrugWars formal runs use a 120-minute per-agent game timeout, a 500-action game cap, an outer 200-turn driver cap, and \texttt{RunConfig(max\_llm\_calls=500)} for each ADK runner invocation.
DrugWars agents must call one exposed game tool per step; if a response contains no tool call, the driver sends a continue prompt, and retryable step errors receive up to three local retries.
Splendor formal runs use a 120-minute arena-round timeout, a 600-second per-decision LLM timeout, at most four LLM calls per player turn, at most eight engine actions per player turn, and a 30-turn cap per player per match.
Splendor turn agents choose from the enumerated legal game actions; history tools are available only in the pre-game history-reading stage and not during turns.
When a Splendor response times out, cannot be parsed, or proposes an invalid action after feedback, the driver applies a deterministic fallback action from the current legal-action list.
MLR-Bench uses the Claude Code harness described above and applies 180-minute timeouts to both agent and judge subprocesses.

\paragraph{Public-summary generation.}
History summaries are generated by the model key supplied through \texttt{--history-summary-model}.
The launcher default is \texttt{gpt\_4o}, but the DrugWars RQ4 summary condition explicitly used \texttt{gpt-5.4}.
Summary calls use no tools, temperature 0.2, \texttt{max\_tokens=2500}, and up to three top-level attempts.
Each top-level summary attempt also uses the same five-attempt retry wrapper as the ADK calls.
The summary input is limited to 60{,}000 characters overall and 10{,}000 characters per source section.
The generated summary is the only previous-round material mounted for agents in the summary history mode.

\end{document}